\documentclass[10pt,twocolumn,letterpaper]{article}

\usepackage{iccv}
\usepackage{times}
\usepackage{epsfig}
\usepackage{graphicx}
\usepackage{amsmath}
\usepackage{amssymb}
\usepackage{bbm}
\usepackage{multirow}
\usepackage[ruled]{algorithm2e}

\usepackage{xcolor}
\usepackage{colortbl}
\usepackage{float}
\usepackage{multirow}
\usepackage{theorem}

\usepackage{pifont}
\definecolor{LightYellow}{RGB}{255,242,204}
\definecolor{LightGray}{gray}{0.93}
\newcolumntype{a}{>{\columncolor{LightGray}}c}

\usepackage[pagebackref=true,breaklinks=true,colorlinks,bookmarks=false]{hyperref}

\iccvfinalcopy 

\def\iccvPaperID{9769} 

\ificcvfinal\fi

\newcommand\Tstrut{\rule{0pt}{2ex}}         
\newcommand\Bstrut{\rule[-1ex]{0pt}{0pt}}   

\begin{document}

\hbadness=2000000000
\vbadness=2000000000
\hfuzz=100pt

\setlength{\abovedisplayskip}{0pt}
\setlength{\belowdisplayskip}{0pt}
\setlength{\textfloatsep}{3pt plus 1.0pt minus 1.0pt}
\setlength{\parskip}{0pt}

\title{Confidence Attention and Generalization Enhanced Distillation for\\Continuous Video Domain Adaptation}

\author{Xiyu Wang, Yuecong Xu, Jianfei Yang, Bihan Wen, Alex C. Kot\\
School of Electrical and Electronic Engineering, Nanyang Technological University, Singapore\\
50 Nanyang Avenue, Singapore 639798\\
{\tt\small \{xiyu001, xuyu0014, yang0478, bihan.wen, eackot\}@ntu.edu.sg}
}

\maketitle
\ificcvfinal\thispagestyle{empty}\fi

\begin{abstract}
Continuous Video Domain Adaptation (CVDA) is a scenario where a source model is required to adapt to a series of individually available changing target domains continuously without source data or target supervision. It has wide applications, such as robotic vision and autonomous driving. The main underlying challenge of CVDA is to learn helpful information only from the unsupervised target data while avoiding forgetting previously learned knowledge catastrophically, which is out of the capability of previous Video-based Unsupervised Domain Adaptation methods. Therefore, we propose a Confidence-Attentive network with geneRalization enhanced self-knowledge disTillation (CART) to address the challenge in CVDA. Firstly, to learn from unsupervised domains, we propose to learn from pseudo labels. However, in continuous adaptation, prediction errors can accumulate rapidly in pseudo labels, and CART effectively tackles this problem with two key modules. Specifically, The first module generates refined pseudo labels using model predictions and deploys a novel attentive learning strategy. The second module compares the outputs of augmented data from the current model to the outputs of weakly augmented data from the source model, forming a novel consistency regularization on the model to alleviate the accumulation of prediction errors. Extensive experiments suggest that the CVDA performance of CART outperforms existing methods by a considerable margin.

\end{abstract}

\section{Introduction}
\begin{figure}[t]
\centering
\includegraphics[width=1.0\linewidth]{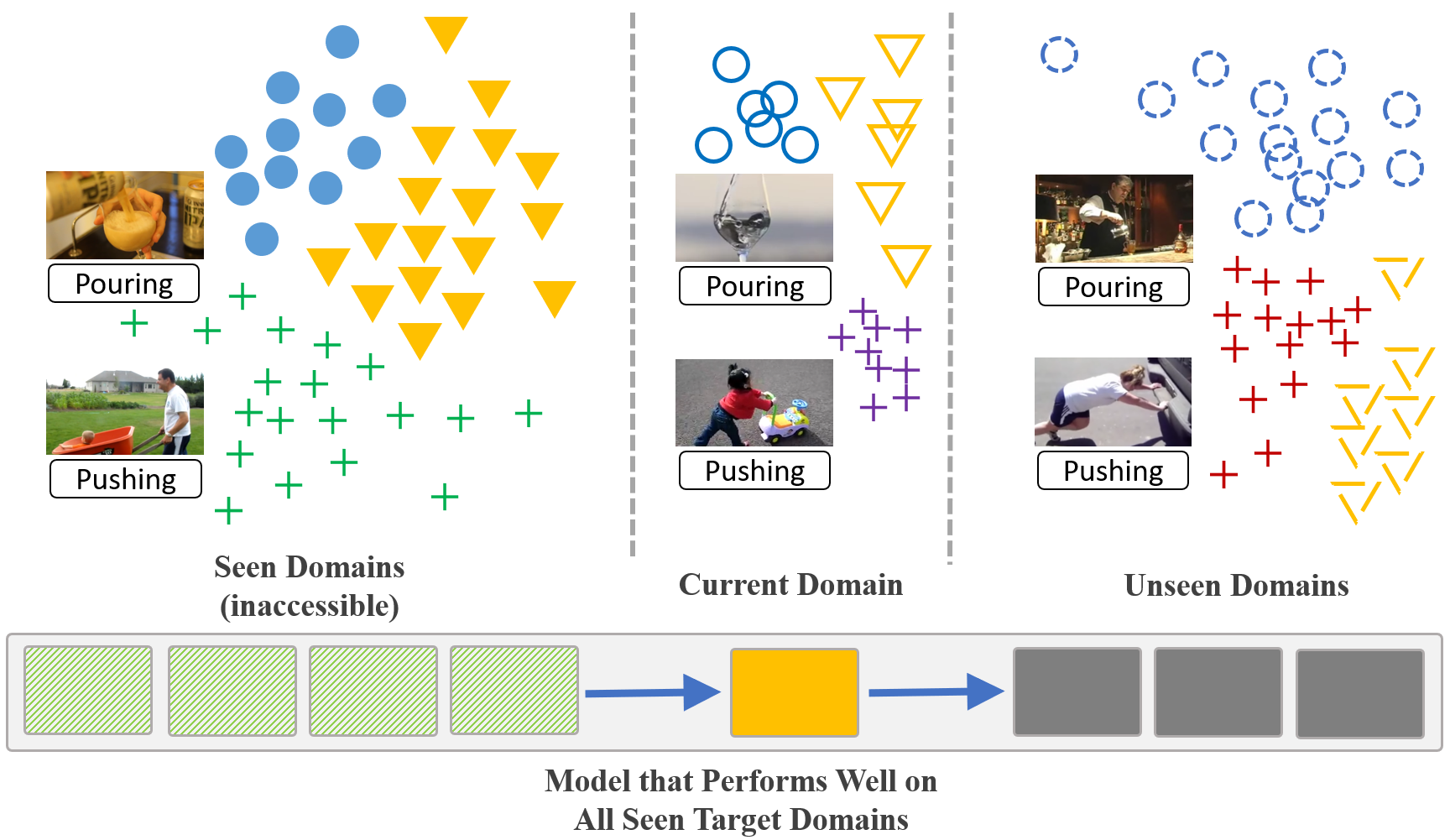}
\caption{Illustration of how changing target domains are encountered in CVDA. The source model is adapted to each individual arriving domain, and seen domains are not accessible. In such a scenario, VUDA methods can forget previously learned knowledge, and it can be limited by storage issues. Therefore, it is nontrivial to tackle the continuous learning challenge in CVDA, and CART is proposed to address these challenges.}
\label{teaser}
\end{figure}

Video action recognition using deep learning~\cite{TSN,C3D,X3D,timesformer} is a long-studied topic and has many applications~\cite{surveillance01,autodrive01,robot01,robot02,device-free}. Still, in real-world scenarios, model adaptation is often necessary due to domain shifts between the source and target domains. Therefore, Video-based Unsupervised Domain Adaptation (VUDA) methods~\cite{VUDA02,VUDA03,VUDA04,TA3N} were proposed to transfer models from a pre-training (source) domain to a static new (target) domain. 

However, continuous adaptation is more desirable for real-world machine perception systems because they are running in continually changing environments, and the target domain distribution can change over time. Intuitively, assuming the source data can be stored and accessed, VUDA methods can continuously adapt to new arriving target domains. Still, this is sub-optimal as VUDA methods do not consider preserving knowledge learned on previous domains, making repeated adaptation necessary when seen domains are re-appearing. On the other hand, there is a strong motivation to enable continuous adaptation without any previous data, including the source data, because storing the massive source data and the accumulated target data for VUDA can be infeasible on platforms such as robots. For example, a public safety robot can encounter videos shot in changing environments (domains), e.g., day and night, sunny and foggy weather, even at the same location, and its onboard storage space can be very limited. To this end, we summarize the aforementioned scenario as Continuous Video Domain Adaptation (CVDA), and the objective for CVDA methods is to obtain a video model that performs well on all seen domains. Specifically, to enable video models to continuously adapt to encountering target domains with only the unsupervised target data from the encountering domain, two goals need to be achieved: 1) the method should enable the model to learn helpful information from each encountering unsupervised target domain without source and previous target domain data, and 2) the method should retain the model performance on all seen target domains.

\begin{table*}[t]
\centering
\label{table_1}
\resizebox{0.75\linewidth}{!}{
	\begin{tabular}{c|c|c|c|c}
		\hline
		Setups & Source Data & Target Supervision & Target Task & Data Type\\ \hline
		UDA~\cite{DANN,MCD,MMD-MK,MDD} & Yes & No & Static Domain& Image\\ 
		VUDA~\cite{TA3N,VUDA02,VUDA03,VUDA04} & Yes & No & Static Domain& Video\\
		SFDA~\cite{SHOT,BAIT,CPGA} & No & No & Static Domain& Image\\
		SFVDA~\cite{SFVDA} & No & No & Static Domain & Video\\
		CiDA~\cite{CiDA} & No & No & Static Class-Incremental & Image\\
		CL~\cite{LwF,EWC,TCD} & N/A & Yes & Continuous Class-Incremental & Image/Video\\
		CDA~\cite{CDA} & N/A & Yes & Continuous Changing Domain & Image\\
		\hline
		\textbf{Ours (CVDA)} & \textbf{No} &  \textbf{No} & \textbf{Continuous Changing Domain} & \textbf{Video}\\ \hline
	\end{tabular}
}
\smallskip
\caption{Differences between CVDA and previous similar setups}
\vspace{-6pt}
\end{table*}

While VUDA methods are failing to achieve those two goals simultaneously, a common strategy to enable unsupervised learning without source data is pseudo labelling~\cite{SHOT,BAIT,CPGA}. These methods focus on adapting to a static target domain and do not consider the accumulation of prediction errors in long-term continual adaptation scenarios. Over time, the accumulated error can significantly mislead the model and cause catastrophic forgetting. Meanwhile, works such as CDA~\cite{CDA} and most Continual Learning (CL) methods consider mitigating catastrophic forgetting and assume the target supervision exists. This is sub-optimal for videos as obtaining target supervision is time and labor-intensive. We list the limitations of existing works in Table~\ref{table_1}. Overall, without effective measurements to simultaneously tame the accumulation of prediction errors over time and learn from unsupervised target domains without source data, previous methods are generally incapable of performing well in the CVDA scenario.

Based on those two aforementioned goals, we identify that the CVDA scenario can be tackled by: \textit{1) robust source-free learning of unsupervised changing target domains, and 2) mitigation of accumulated prediction errors that mislead models and cause catastrophic forgetting}. In view of this, we propose \textbf{Confidence-Attentive network with geneRalization enhanced self-knowledge disTillation (CART)}. Firstly, CART uses target predictions as pseudo labels and deploys prototypical classification to refine noisy pseudo labels. To reduce the unreliability of pseudo labels when the target domain changes continuously, we propose a new learning strategy where samples are learned attentively based on their prediction confidence such that high confidence samples are learned to lead the model update while low confidence but correct samples can also contribute to the acquirement of new knowledge. 
Secondly, observing that the source model is often less biased and more plastic than target models trained with noisy pseudo labels, we propose to regularize the current model parameter to behave similarly to the source model. This is achieved by constructing a self-knowledge distillation process enhanced by data generalization. Specifically, the current model is first fed with generalized, i.e., strongly augmented, data, and the source model is fed with weakly augmented data. Next, the current model is enforced to mimic the output of the source model to ensure the current model behaves similarly to the source model. This ensures that the current model behavior is consistent with the less biased source model in a larger domain, i.e., the generalized target domain, and the accumulation of prediction errors is better reduced. Extensive experiments show that CART can effectively reduce the accumulation of prediction errors and achieves state-of-the-art continuous adaptation results on new dedicated CVDA benchmarks.

In summary, this paper has made the following contributions: 1) to the best of our knowledge, this paper is the first research to touch on the concept of CVDA, where deep video models are required to continuously adapt to new target domains while retaining learned knowledge on seen domains without the source data and target supervision; 2) we analyze the challenges underlying CVDA and introduce CART to address these challenges by attentive learning of pseudo labels and generalization enhanced self-knowledge distillation; and 3) we construct two dedicated CVDA benchmarks, namely Sports-DA\textsubscript{\it Conti.} and Daily-DA\textsubscript{\it Conti.}. Extensive experiments demonstrate that CART achieves an average of 8.42\% relative improvement in adaptation performances over previous state-of-the-art methods. Meanwhile, CART also has an average forgetting rate of -1.09\%, indicating that CART enables knowledge accumulation instead of forgetting.

\section{Related Works}

\noindent \textbf{Video-based Unsupervised Domain Adaptation}
Unsupervised Domain Adaptation (UDA) methods aim to transfer source domain image knowledge to an unsupervised target image domain. The mainstream idea is to mitigate domain discrepancy~\cite{MCD,MDD,MMD-MK,DANN}. Recent UDA studies propose other ideas, such as varying the input space~\cite{cycda} or leveraging self-training~\cite{self-training}. In contrast, recent Video-based Unsupervised Domain Adaptation (VUDA) methods~\cite{TA3N,SFVDA,VUDA02,VUDA03,VUDA04} focus on enabling efficient transfer of video models. The main underlying challenge in VUDA is that generating domain-invariant video features is more difficult due to the complexity added by the temporal dimension. As a result, recent VUDA works~\cite{TA3N,VUDA02,VUDA03,VUDA04} primarily focus on additionally aligning temporal features. Recently, researchers are gradually shifting their focus from the standard domain adaptation of videos to other related domains, such as source-free video domain adaptation~\cite{SFVDA} and multi-source video domain adaptation~\cite{MSVDA}.

\noindent \textbf{Continual Learning}
The main challenge in Continual Learning (CL) is to prevent catastrophic forgetting when learning new tasks~\cite{CL-Defy}. Continual learning methods can be divided into two categories: replay-based~\cite{icarl} and regularization-based~\cite{LwF,LFL,MAS}. Replay-based methods memorize a certain amount of source data to prevent forgetting, while regularization-based methods~\cite{LwF,LFL,MAS} seek to regularize the model update without any source data to mitigate forgetting. Among many CL methods, Knowledge Distillation~\cite{KD} (KD) is particularly effective and does not require accessing source data. It was originally used for knowledge transfer from big to small models and was considered an effective CL method~\cite{LwF,LFL,CI-Consolidation}. In this paper, we found it offers simple but effective regularization on the current model in terms of preventing the accumulation of prediction errors.

\noindent \textbf{Continuous Domain Adaptation}
While most CL methods are developed for class-incremental scenarios, others focus on continuous domain adaptation. Early works such as CMA~\cite{CMA} and IADA~\cite{IADA} addressed the problem of adapting to evolving target image data. More recently, CDA~\cite{CDA} proposed the use of meta-learning to adapt to a series of supervised target domains. Studies such as CiDA~\cite{CiDA} instead combine domain adaptation with class-incremental continual learning. In other pioneering works~\cite{CoTTA}, the combined problem of test-time adaptation and continuous image domain adaptation was tackled. 

\noindent \textbf{Domain Generalization.}
Domain generalization is a well-studied problem in computer vision, and many related methods have been proposed~\cite{dg1,dg2,dg3}. Among these methods, data augmentation is particularly effective. Image-transform-based methods~\cite{dg4,dg5} have shown that image transformations can generalize source data to out-of-distribution data, enabling better adaptation performance. More recent studies propose to leverage neural networks to manipulate the augmentation process of images~\cite{dg6,dg7}. For a better understanding of domain generalization, we refer readers to related surveys~\cite{dg8}. In this paper, we show that simple domain generalization techniques can further improve the effectiveness of self-knowledge distillation.

\section{Continuous Video Domain Adaptation}
\subsection{Problem Definition}
Given an pre-trained model consisting of a feature extractor $g_0$ with parameters $\theta_0$ and classifier $h_0$ with parameters $\phi_0$ trained on the source data $(\mathcal{X}^S,\mathcal{Y}^S)$,  we aim at adapting to a series of target domains $\{\mathcal{D}_1, \mathcal{D}_2, \mathcal{D}_3,...\mathcal{D}_t,...\}$ where $\mathcal{D}_t=\{x_{it}\}^{N_{t}}_{i=1}$ with $N_{t}$ i.i.d. videos $x_{it}$. The adaptation starts from the $t=1$ moment, i.e., the source domain is not accessible, and $\mathcal{D}_t$ is only available at the current time step. We assume the label spaces $C_t$ across all domains are identical, and the source task $\mathcal{X}^0\rightarrow\mathcal{Y}^0$ is the same as all target tasks $\mathcal{X}^{t}\rightarrow\mathcal{Y}^{t}$. In this manner, CVDA focuses on two metrics: 1) adaptation performance on $\mathcal{D}_t$ and 2) prediction performance on all seen target domains, e.g., $\mathcal{D}_{1}, \mathcal{D}_{2}, ..., \mathcal{D}_{t-1}$. We assume $t\ge1$ by default. Thus, without further notification, when $t$ appears in a formula in this paper, it refers to a time step greater than 0. For simplicity, $i$ in $x_{it}$ is omitted in some formulas, and $x_t$ is weakly augmented by default.

The aforementioned CVDA setup is designed to address the growing demand for continuous adaptation to new video domains with limited data in many real-world applications. In CVDA, deep models cannot rely on source data and target supervision and are prone to noisy inputs. Moreover, CVDA also requires consistent performance on all previously seen target domains, indicating that methods designed for CVDA should preserve previous domain knowledge and further enable the accumulation of knowledge across domains to improve the adaptation performance on the newly learned domain.

\subsection{Methodology \label{method}}

\noindent \textbf{Source Model Generation.} Many existing Source-Free Domain Adaptation (SFDA) methods~\cite{SHOT,DINE,CiDA} require a specific source model preparation phase where modified training strategies, such as entropy loss~\cite{SHOT,TENT}, are applied. In contrast, CART does not require specific source preparation. Moreover, besides proposing CVDA, this paper is also a pioneering video domain adaptation work that is solely based on transformer-based networks. We use TimeSFormer~\cite{timesformer} as the backbone for all involved methods in this paper, for it is relatively lightweight, performative, and robust.

\begin{figure*}[!ht]
\centering
\includegraphics[width=.85\linewidth]{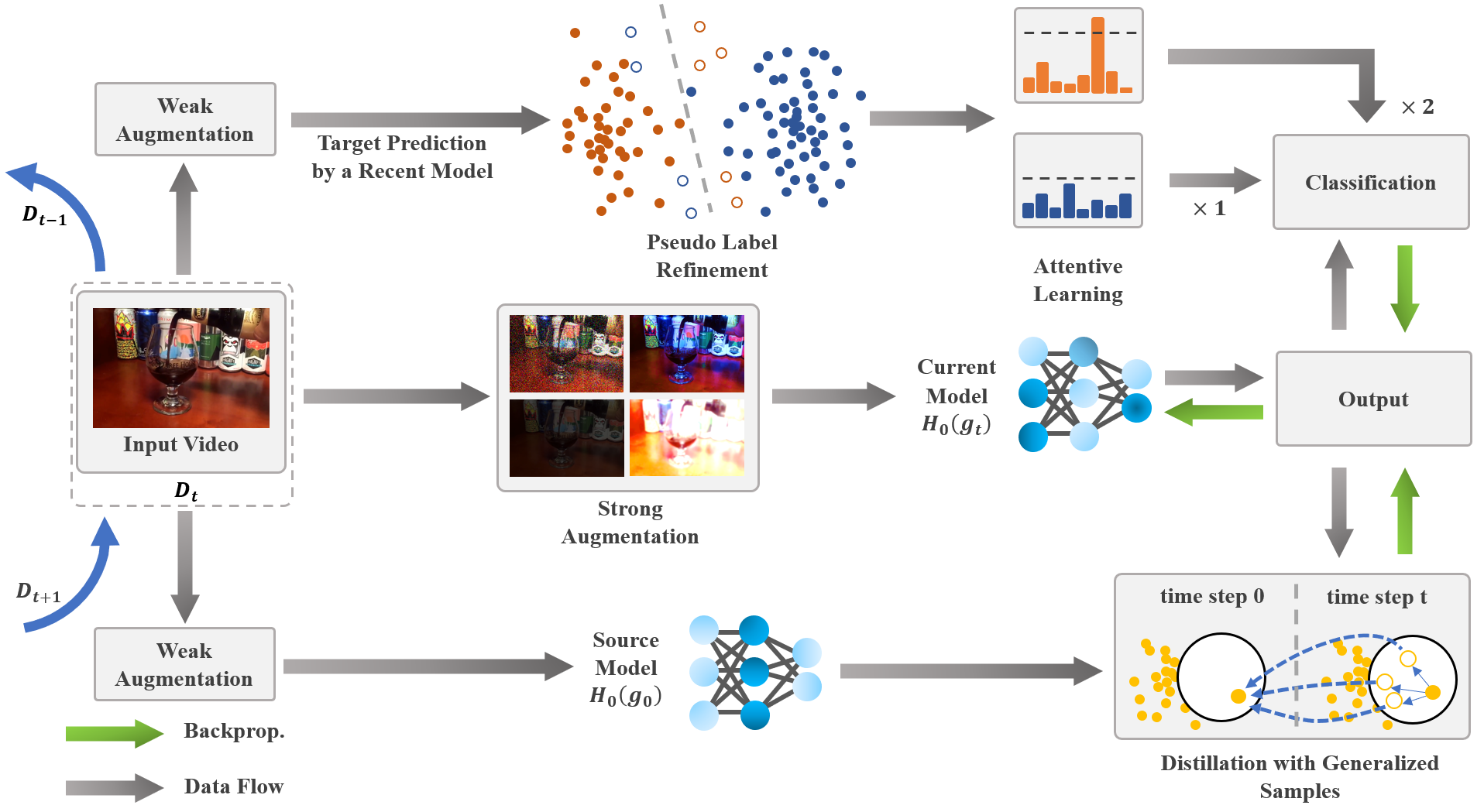}
\caption{Graphical illustration of the proposed CART. Whenever a new domain $\mathcal{D}_t$ is met at time step $t$, CART first feeds weakly augmented samples to the source model $h_0(g_0)$ and a previous version of $h_0(g_t)$ saved recently. Next, CART feeds strongly augmented samples to the current model $h_0(g_t)$. With the model output, the classification loss $\mathcal{L}_{Acls}$ is obtained using Eqn.~\ref{new_ce_loss}. Combining with the distillation loss $\mathcal{L}_{Adis}$, we obtain the loss to be optimized in CART, i.e., Eqn.~\ref{main_loss}. With the combined loss, we update the current model $h_0(g_t)$ with stochastic gradient descent until epochs on the data batch for $\mathcal{D}_t$ is completed. \label{Method_Fig}} 
\end{figure*}

\noindent \textbf{Attentive Learning with Pseudo Labels.}\label{Pseudo_labeling}
Modern deep classification models are commonly constructed as two parts, i.e., a feature extractor $g_0$ and a classifier $h_0$. UDA methods~\cite{DANN,MCD,MMD-MK,MDD} often aim to generate domain invariant features by minimizing metrics that could reflect the degree of domain discrepancy. Thus, the loss to be optimized for a single source-target sample pair of samples in UDA methods can be summarized as follows:
\begin{equation}
\mathcal{L}=\mathcal{L}_{cls}(x_s,y_s)+\mathcal{L}_{dom}(x_s,x_t),
\end{equation}
where $\mathcal{L}_{cls}$ is the source classification loss,  $\mathcal{L}_{dom}$ is a domain loss that takes both source and target data as input, $x_s$ is a source sample. This is not applicable in the proposed CVDA setup where source data are not accessible. In view of this, SFDA methods~\cite{SHOT,BAIT,CPGA} propose either learning from pseudo labels or optimizing other types of unsupervised losses~\cite{SHOT}. In this paper, inspired by previous studies~\cite{SHOT,TENT}, we take a simple yet effective way to obtain refined pseudo labels. Formally, given target features $g_t(x_t)$ and target predictions $h_t(g(x_t))$, the centroids of target features can be obtained by the weighted average of all features on each class as follows:
\begin{equation}
\label{SHOT_centroids}
c_k=\frac{\sum_{x_t\in\mathcal{X}^t}\delta_k(h_0(g_t(x_t))g_t(x_t))}{\sum_{x_t\in\mathcal{X}^t}\delta_k(h_0(g_t(x_t)))},
\end{equation}
where $\delta_k$ is a softmax function and $k$ indexes the $k$-th class. Then, by classifying each feature to the corresponding class of its closest centroid, the refined pseudo label is:
\begin{equation}
\label{get_SHOT_label}
\hat{y_t}=\text{arg}\operatorname*{\min}_{k}Dist(g_t(x_t),c_k),
\end{equation}
where $Dist$ measures the cosine distance between a target feature to all centroids, notice that we do not optimize the parameter $\phi_0$ of the source classifier $h_0$ as previous works suggest that this can be unnecessary~\cite{SHOT}.

Refined pseudo labels can still contain a considerable amount of prediction errors when the domain shift is large. In such scenarios, even state-of-the-art refinement strategies can only provide minimal improvement, making regularizing the model to reduce the accumulation of prediction errors necessary. In view of this, we propose to learn from samples attentively. Specifically, we propose to learn from each prediction attentively based on their prediction confidence. Firstly, the prediction confidence of sample $x_t$ is calculated as follows:
\begin{equation}
Conf(x_t)=max(\delta(h_0(g_t(x_t)))).
\end{equation}
While previous methods~\cite{fixmatch} forsake all low-confidence examples, we find this is wasteful as there can be many low-confidence correct predictions in CVDA. Therefore, we instead propose to pay attention to high-confidence and low-confidence samples differently such that the former can lead the optimization while new knowledge can also be learned from the latter, i.e., 
\begin{equation}
\label{new_ce_loss}
\resizebox{0.85\linewidth}{!}{$
	\begin{aligned}
		\mathcal{L}_{cls}=\frac{1}{N_t}\sum_{i=1}^{N_t}\sum_{m=1}^M(\mathbbm{1}(Conf(x_{it})>\tau_m)\mathcal{L}_{ce}(x_{it}, \hat{y_{it}})),
	\end{aligned}
	$}
\end{equation}
where $\mathbbm{1}(Conf(x_{it})>\tau_m)$ masks out predictions whose confidence are lower than $\tau_m$, and $\mathcal{L}_{ce}$ is a standard cross-entropy loss. In practice, we find setting $M=2,\tau_1=0.1,\tau_2=0.5$ generally sufficient.

\noindent \textbf{Self-Knowledge Distillation with Generalized Samples.} Though learning attentively from refined pseudo labels is effective, it cannot guarantee that the accumulation of prediction errors is reduced. This is because pseudo labels inevitably contain errors, and the error rate is likely to be reinforced when continuously adapting for a long time. What's worse is the domain shifts between video domains are often large, making the error rate grow rapidly as the model is misled continuously by noisy pseudo labels. Eventually, without the help of source and target supervision, the model may not be able to recover from such a state by itself. In view of this, inspired by some CL methods~\cite{LwF,LFL}, we propose to regularize the model update such that $h_0(g_{t})$ is enforced to behave similarly as $h_0(g_{0})$ via self-knowledge distillation. By enforcing the current model $h_0(g_{t})$ to behave similarly to the source model $h_0(g_{0})$, the accumulation of errors can be kept at a lower level such that the model can retain learned knowledge. Formally, we can obtain the self-distillation loss as follows:
\begin{equation}
\label{KL_div}
\resizebox{.8\linewidth}{!}{$
	\begin{aligned}
		\mathcal{L}_{dis}&=\frac{1}{N_t}\sum_{i=1}^{N_t}\text{KL}(\delta(h_0(g_0(x_{it})))|\delta(h_0(g_t(x_{it}))))\\
		&= \frac{1}{N_t}\sum_{i=1}^{N_t} \delta(h_0(g_0(x_{it}))) \log \frac{\delta(h_0(g_0(x_{it})))}{\delta(h_0(g_t(x_{it})))},
	\end{aligned}
	$}
\end{equation}
where the source model $h_0(g_0)$ is regarded as a teacher model, and the model at time step $t$ is considered as a student model. By minimizing $\mathcal{L}_{dis}$, the current model $h_0(g_{t})$ can receive knowledge from the less biased $h_0(g_{0})$ and keep itself less biased.

Notably, we denote the distillation loss that uses the model from $t-1$ moment as the teacher model $\mathcal{L}_{dis}'$. This is a common practice in previous works~\cite{LwF}, but it is unsuitable in CVDA since 1) frequently saving $g_{t-1}$ is space inefficient, and 2) models trained with pseudo labels are generally biased by accumulated prediction errors. 

With the aforementioned process, we observe that the accumulation of prediction errors is reduced. Still, we find the results are sometimes unsatisfactory when we want the model to learn new knowledge while having minimal forgetting rates. Intuitively, enlarging the trade-off for $\mathcal{L}_{dis}$ can ensure a better reduction of the accumulation of prediction errors. However, we observe that large trade-offs often limit the model to learn new knowledge. To achieve stronger regularization without damaging model plasticity, i.e., the ability to learn new knowledge, we propose to enhance the regularization on $h_0(g_t)$ by first strongly augmenting samples fed to it and then compute the distillation loss between the output of strongly augmented samples from $h_0(g_t)$ and the output of weakly augmented samples from $h_0(g_0)$. To this end, the overall training and evaluation pipeline is also illustrated in Algorithm \ref{algo_1}, and we re-write Eqn.~\ref{new_ce_loss} and Eqn.~\ref{KL_div} to obtain the loss that is optimized in CART:
\begin{equation}
\label{main_loss}
\resizebox{0.85\linewidth}{!}{$
	\begin{aligned}
		\mathcal{L}&=\mathcal{L}_{Acls}+\mathcal{L}_{Adis}\\
		&=\frac{1}{N_t}\sum_{i=1}^{N_t}\sum_{m=1}^M(\mathbbm{1}(Conf(\mathcal{A}(x_{it}))>\tau_m)\mathcal{L}_{ce}(\mathcal{A}(x_{it}), \hat{y_{it}}))\\
		&+\frac{\alpha}{N_t}\sum_{i=1}^{N_t} \delta(h_0(g_0(x_{it}))) \log \frac{\delta(h_0(g_0(x_{it})))}{\delta(h_0(g_t(\mathcal{A}(x_{it}))))},
	\end{aligned}
	$}
\end{equation}
where $\mathcal{A}$ represents the strong augmentation applied to the input data $x_{it}$, $\mathcal{L}_{Acls}$ and $\mathcal{L}_{Adis}$ are classification and distillation loss that receives strongly augmented samples as their input from $h_0(g_t)$. The key motivation behind constructing $\mathcal{L}_{Adis}$ in Eqn.~\ref{main_loss} is that, instead of only regularizing the model to respond similarly when the input is $x_t$, we instead regularize the model to respond similarly to the source model output $h_0(g_0(x_t))$ when the generalized version of the input, e.g., $\mathcal{A}(x_t)$, arrives. This enforces the model to act similarly to the source model not only in the target domain $\mathcal{D}_{t}$ but also on a generalized target domain $\hat{\mathcal{D}_{t}}$. Empirically, we find this strategy more effective than distilling with $\mathcal{L}_{dis}$, indicating that generalization can allow better model plasticity while retaining previous knowledge. Notably, we also apply strong augmentation to $\mathcal{L}_{cls}$, i.e., $\mathcal{L}_{Acls}$, forming a structure similar to some previous works~\cite{fixmatch}. Such structure for classification is empirically proven to be effective by \cite{fixmatch} in terms of improving adaptation performance, and it also simplifies the data processing pipeline of CART shown in Fig.~\ref{Method_Fig}.

\begin{algorithm}[!t]
\caption{The training pipeline of CART \label{algo_1}}
\small
\KwData{Streams of target data $x_t$ on a target domain sequence $\{\mathcal{D}_1, \mathcal{D}_2, \mathcal{D}_3,...\}$, Source model $h_0(g_0)$.}
\KwResult{Model $h_0(g_t)$ adapted to $\mathcal{D}_t$}
\For{ Epochs on $\mathcal{D}_t$ in $\{\mathcal{D}_1, \mathcal{D}_2, \mathcal{D}_3,...\}$}{
	Weakly augment $x_t$\;
	Obtain all source model outputs $h_0(g_0(x_t))$\;
	Obtain all $\hat{y_t}$ via Eqn.~\ref{get_SHOT_label} every few epochs\;
	\For{$x_t$ in $\mathcal{D}_t$}{
		Strongly Augment $x_t$ as $A(x_t)$\;
		Obtain classification and distillation loss via Eqn.~\ref{main_loss}\;
		Back propagation\;
	}
	Update parameter $\theta_{t-1}\rightarrow\theta_{t}$\;
	Evaluation on $\mathcal{D}_t,\mathcal{D}_{t-1},...,\mathcal{D}_1$\;
}

\end{algorithm}

\section{Experiments}
We conduct experiments on new benchmarks designed for Continuous Video Domain Adaptation (CVDA). We compare our method against other representative methods and demonstrate that our method can efficiently tackle the challenge of CVDA. We further present in-depth analysises of modules in CART to justify the design of CART and empirical analysis to demonstrate the effectiveness of CART. \textit{Extra implementation details are in the appendix.}

\begin{table*}[t]
\begin{center}
	\resizebox{0.73\linewidth}{!}{
		\begin{tabular}{ccccccc|c|c}
			\hline\hline
			\multicolumn{9}{c}{time $\xrightarrow{\hspace*{15cm}}$}\\
			\hline
			\multicolumn{1}{c|}{method}      & \parbox{1.2cm}{\centering ARID\textsubscript{1}} & \parbox{1.2cm}{\centering MIT\textsubscript{1}} & \parbox{1.2cm}{\centering HMDB\textsubscript{1}} & \parbox{1.2cm}{\centering ARID\textsubscript{2}} & \parbox{1.2cm}{\centering MIT\textsubscript{2}} & \parbox{1.2cm}{\centering HMDB\textsubscript{2}} & Mean Acc. & Mean Forget\\ \hline
			\multicolumn{1}{c|}{Source-Only} & 20.81\%  &  \textbf{26.91}\%  &  40.00\%  &  20.81\%  &  26.91\%  &  40.00\%  &  29.24\%  & N/A \\
			\hline
			\multicolumn{1}{c|}{SHOT}        &  21.55\%  &  24.91\%  &  29.39\%  &  21.95\%  &  25.27\%  &  37.27\%  &  26.72\%  &  -1.59\%\\
			\hline
			\multicolumn{1}{c|}{ACAN}        &  21.67\%  &  25.73\%  &  38.03\%  &  21.48\%  &  24.85\%  &  33.55\%  &  27.55\%  &  7.42\%\\ 
			\multicolumn{1}{c|}{DANN}        &  \textbf{21.78}\%  &  25.27\%  &  37.88\%  &  21.32\%  &  26.00\%  &  39.39\%  &  28.61\%  & -0.81\%\\
			\multicolumn{1}{c|}{MCD}         &  15.33\%  &  23.45\%  &  31.21\%  &  12.10\%  &  19.09\%  &  25.45\%  &  21.11\%  & 1.24\%\\
			\hline
			\multicolumn{1}{c|}{TENT}        &  20.81\%  &  26.18\%  &  36.97\%  &  20.76\%  &  25.27\%  &  36.36\%  &  28.03\%  & 2.78\%\\
			\multicolumn{1}{c|}{CoTTA}       &  20.53\%  &  \textbf{26.91}\% &  39.39\%  &  18.27\%  &  26.73\%  &  36.67\%  & 28.68\%   &  0.66\% \\
			\hline
			\multicolumn{1}{c|}{CART}       &  \textbf{21.78}\%  &  \textbf{26.91}\%  &  \textbf{45.76\%}  &  \textbf{26.19\%}  &  \textbf{27.09\%}  &  \textbf{47.58\%}  &  \textbf{32.55\%}  & \textbf{-3.83\%}\\
			\hline\hline
		\end{tabular}
	}
	\smallskip
	\caption{Domain adaptation results on ARID$\rightarrow$MIT$\rightarrow$HMDB51 \label{table_2}}
\end{center}
\vspace{-14pt}
\end{table*}

\subsection{Experimental Settings}
Our experiments are conducted on two CVDA benchmarks: Daily-DA\textsubscript{\it Conti.}, and Sports-DA\textsubscript{\it Conti.}.

\begin{table*}[!ht]
\begin{center}
	\resizebox{0.73\linewidth}{!}{
		\begin{tabular}{ccccccc|c|c}
			\hline\hline
			\multicolumn{9}{c}{time $\xrightarrow{\hspace*{15cm}}$}\\
			\hline
			\multicolumn{1}{c|}{method}      & \parbox{1.2cm}{\centering HMDB\textsubscript{1}} & \parbox{1.2cm}{\centering ARID\textsubscript{1}} & \parbox{1.2cm}{\centering MIT\textsubscript{1}} & \parbox{1.2cm}{\centering HMDB\textsubscript{2}} & \parbox{1.2cm}{\centering ARID\textsubscript{2}} & \parbox{1.2cm}{\centering MIT\textsubscript{2}} & Mean Acc. & Mean Forget\\ \hline
			\multicolumn{1}{c|}{Source-Only} &  40.00\%  &  20.81\%  &  26.91\%  &  40.00\%  &  20.81\%  &  26.91\%  &  29.24\%  & N/A\\
			\hline
			\multicolumn{1}{c|}{SHOT}        &  43.94\%  &  24.55\%  &  25.64\%  &  42.73\%  &  23.42\%  &  26.73\%  &  31.17\%  &  2.33\%\\
			\hline
			\multicolumn{1}{c|}{ACAN}        &  35.31\%  &  20.84\%  &  27.51\%  &  37.25\%  &  21.81\%  &  25.62\%  &  28.05\%  & 4.05\% \\ 
			\multicolumn{1}{c|}{DANN}        &  32.73\%  &  20.53\%  &  25.27\%  &  35.15\%  &  20.87\%  &  24.91\%  &  26.57\%  & \textbf{-0.51}\%\\
			\multicolumn{1}{c|}{MCD}         &  31.82\%  &  20.59\%  &  22.73\%  &  29.39\%  &  15.84\%  &  19.09\%  &  23.24\%   &  2.13\% \\
			\hline
			\multicolumn{1}{c|}{TENT}        &  36.36\%  &  20.81\%  &  25.82\%  &  36.97\%  &  20.81\%  &  25.27\%  &  25.27\%  & 0.27\%\\
			\multicolumn{1}{c|}{CoTTA}       &  37.88\%  &  20.87\%  &  26.36\%  &  39.39\%  &  20.70\%  &  26.36\%  &  28.59\%  & 1.63\% \\
			\hline
			\multicolumn{1}{c|}{CART}       &  \textbf{46.06\%}  &  \textbf{27.94\%}  &  \textbf{28.73\%}  &  \textbf{49.39\%}  &  \textbf{29.19\%}  &  \textbf{30.18\%}  &  \textbf{35.24\%}  & -0.19\%\\
			\hline\hline
		\end{tabular}
	}
	\smallskip
	\caption{Domain adaptation results on HMDB51$\rightarrow$MIT$\rightarrow$ARID  \label{table_3}}
\end{center}
\vspace{-14pt}
\end{table*}

\textbf{Daily-DA\textsubscript{\it Conti.}} is modified based on Daily-DA proposed by MSVDA~\cite{MSVDA}, it is constructed upon four mainstream datasets, e.g. Kinetics-600(K600)~\cite{K600}, HMDB51(HMDB)~\cite{HMDB51}, ARID~\cite{ARID}, and Moments-in-Time(MIT)~\cite{moments-in-time}. Kinetics-600, HMDB51, and Moments-in-Time are three prevailing large datasets used for video action recognition benchmarking. ARID is a dataset specifically built for action recognition in dark environments, and its data distribution drastically differs from others, making it challenging for video models to perform well on it. 11 classes are collected in this benchmark, and more than 20000 videos are available. Due to Daily-DA being a multi-source domain adaptation benchmark, we made the following modifications to enable it as a benchmark for CVDA: 1) Kinetics-600 is regarded as a fixed source domain, and 2) others are split into two non-overlapping parts equally. The motivation behind the dataset split is that we intend to train our methods on a prolonged sequence of domains so that the benchmark can better simulate real-world scenarios where changing target domains arrives continuously and sometimes seen domains can re-appear. As a result, HMDB51 is split into HMDB\textsubscript{1} and HMDB\textsubscript{2}. Splits for Kinetics600 and MIT also follow the same naming convention. With training data prepared, we mainly experiment on two sequences, e.g., ARID\textsubscript{1}$\rightarrow$MIT\textsubscript{1}$\rightarrow$HMDB\textsubscript{1}$\rightarrow$ARID\textsubscript{2}$\rightarrow$MIT\textsubscript{2}$\rightarrow$HMDB\textsubscript{2} and HMDB\textsubscript{1}$\rightarrow$ARID\textsubscript{1}$\rightarrow$MIT\textsubscript{1}$\rightarrow$HMDB\textsubscript{2}$\rightarrow$ARID\textsubscript{2}$\rightarrow$MIT\textsubscript{2}. For simplicity, we refer to these two sequences as ARID$\rightarrow$MIT$\rightarrow$HMDB51 and HMDB51$\rightarrow$MIT$\rightarrow$ARID, respectively in following tables, figures, and paragraphs.

\textbf{Sports-DA\textsubscript{\it Conti.}} is processed similar to Daily-DA. It is also an adapted version of the original Sports-DA proposed by MSVDA~\cite{MSVDA}. This benchmark contains data from three mainstream datasets, e.g., Sports-1M~\cite{Sports1M}, Kinetics-600~\cite{K600}, and UCF101~\cite{UCF101}. 23 classes were involved, and over 40000 videos are available for benchmarking. Similarly, regarding UCF101 as the fixed source dataset, we split the training data of Kinetics-600 and Sports-1M into two non-overlapping splits equally to create four changing batches of data to adapt to, e.g., K600\textsubscript{1}, K600\textsubscript{2}, Sports1M\textsubscript{1}, and Sports1M\textsubscript{2}. The sequence we use during experimenting is Sports1M\textsubscript{1}$\rightarrow$K600\textsubscript{1}$\rightarrow$Sports1M\textsubscript{2}$\rightarrow$K600\textsubscript{2}. Similarly, we refer to this sequence as Sports1M$\rightarrow$Kinetics600 for simplicity.

We implement all methods using the PyTorch~\cite{pytorch} library. We use TimeSFormer~\cite{timesformer} as the backbone. UDA/VUDA~\cite{DANN,MCD,ACAN}, SFDA~\cite{SHOT}, and TTA~\cite{TENT,CoTTA} methods are involved as baselines. For UDA/VUDA methods, we made source data available to them, while SFDA and TTA methods are directly employed. Additionally, for a fair comparison, TTA methods are evaluated on the mixture of train and test split, and the reported result is the accuracy of the test split. We generate our source model based on weights trained on Kinetics-400~\cite{K400}. We use the SGD optimizer with a default learning rate of 0.001. All involved methods are trained for 10 epochs on each domain before the next domain arrives. The learning rate is decreased to 0.0001 at epoch 5 for UDA/VUDA methods, SFDA methods, and CART. Pseudo labels are generated based on task predictions using a previous version of $h_0(g_t)$ that is saved every 5 epochs. We set $\alpha=5$ in Eqn.~\ref{main_loss} as it offers a reasonable level of regularization without damaging model plasticity. For UDA/VUDA methods~\cite{DANN,MCD,ACAN}, the source data is made available to them. As for the weak augmentation and strong augmentation, they are similar to the augmentation set introduced in FixMatch~\cite{fixmatch} while the strong augmentation set is tuned to be more aggressive.

\subsection{Results and Comparisons}

\begin{table}[!ht]
\centering
\resizebox{1\linewidth}{!}{
	\begin{tabular}{ccccc|c|c}
		\hline\hline
		\multicolumn{7}{c}{time $\xrightarrow{\hspace*{12cm}}$}\\
		\hline
		\multicolumn{1}{c|}{method}      
		& \parbox{1.2cm}{\centering Sports1M\textsubscript{1}} 
		& \parbox{1.2cm}{\centering K600\textsubscript{1}} 
		& \parbox{1.2cm}{\centering Sports1M\textsubscript{1}} 
		& \parbox{1.2cm}{\centering K600\textsubscript{2}} & Mean Acc. & Mean Forget\\ \hline
		\multicolumn{1}{c|}{Source-Only} &  80.89\%  &  89.91\%  & 80.89\%  &  89.91\% & 85.40\% & N/A\\
		\hline
		\multicolumn{1}{c|}{SHOT}        &  82.37\%  &  93.32\%  &  82.00\%  &  93.43\%  & 87.78\% & 1.06\% \\
		\hline
		\multicolumn{1}{c|}{ACAN}        &  81.21\%  &  91.36\%  & 82.44\%   &  91.36\%  & 86.59\% & 2.52\%\\ 
		\multicolumn{1}{c|}{DANN}        &  80.16\%  &  91.08\%  &  81.37\%  &  91.03\%  & 85.91\% & -0.34\% \\
		\multicolumn{1}{c|}{MCD}         &  79.42\%  &  88.43\%  &  78.63\%  &  85.58\%  & 83.02\% & 0.51\%\\
		\hline
		\multicolumn{1}{c|}{TENT}        &  79.37\%  &  90.27\%  &  80.63\%  & 90.52\%   &  85.20\% & \textbf{-0.54}\%\\
		\multicolumn{1}{c|}{CoTTA}       &  79.63\%  &  87.67\%  &  75.47\%  &  79.82\%  & 80.39\% & 5.63\% \\
		\hline
		\multicolumn{1}{c|}{CART}       &  \textbf{82.53\%}  &  \textbf{94.80\%}  &  \textbf{82.58\%}  &  \textbf{94.34\%}  & \textbf{88.56\%} & 0.73\% \\
		\hline\hline
	\end{tabular}
}
\smallskip
\caption{Domain adaptation results on Sports1M$\rightarrow$Kinetics600 \label{table_4}}
\end{table}

In our study, we evaluate the performance of CART against several existing state-of-the-art methods, including SFDA methods such as SHOT~\cite{SHOT}, UDA/VUDA methods including DANN~\cite{DANN}, MCD~\cite{MCD}, and ACAN~\cite{ACAN}, and test-time adaptation methods TENT~\cite{TENT} and CoTTA~\cite{CoTTA}. Negative learning rates indicate the model is learning instead of forgetting knowledge. The average forgetting rate is calculated between $h_0(g_t)$ and $h_0(g_{t-1})$ on all seen domains. Notably, average forgetting rates can be numerically small, but it can quickly lead to significant performance regression as time step $t$ increases.

Results in Table~\ref{table_2}-\ref{table_4} show that the novel CART achieves state-of-the-art adaptation results consistently on three challenging benchmarks, outperforming the best-performing prior UDA/VUDA, SFDA, and TTA methods by an average improvement of  11.32\% on HMDB51$\rightarrow$MIT$\rightarrow$ARID, 13.06\% on HMDB51$\rightarrow$ARID$\rightarrow$MIT, and 0.89\% on Sports1M$\rightarrow$Kinetics600. We identify that previous UDA/VUDA methods~\cite{DANN,MCD,ACAN} are surpassed by our methods by a large margin in terms of both average adaptation performances and sometimes the average forgetting rates, indicating that CART can remain stable even without the support of source data. Specifically, CART is obtaining the lowest average forgetting rate in Table~\ref{table_2} while being competitive in Table~\ref{table_3},\ref{table_4} when compared to UDA/VUDA methods that can access the source data, meaning that accumulation of prediction error is effectively mitigated even without the help of source data. With the low forgetting rate and high adaptation performance, the empirical results fully justify that CART has effectively tackled the challenge brought by CVDA and enables the continuous learning of a global model.

Furthermore, it is observed that both SFDA methods~\cite{SHOT} and UDA/VUDA~\cite{DANN,MCD,ACAN} are achieving unsatisfactory results and sometimes worse than the source model. For SFDA methods, e.g., SHOT~\cite{SHOT}, we identify that they are particularly prone to extremely noisy target domains, such as the ARID domain. Without measurements to reduce the accumulation of prediction errors, the continuous adaptation performance of SHOT\cite{SHOT} drops constantly and heavily. We also find that, even with source data being accessible, UDA/VUDA methods are still incapable of adapting to continuously changing domains. We believe such results are caused by: 1) the Daily-DA\textsubscript{\it Conti.} is very challenging, and the domain adversarial network in UDA/VUDA can be unstable when the target domain is frequently varying. Finally, we find that TTA methods~\cite{TENT,CoTTA} are also achieving degraded results. We believe that this is because TTA methods cannot utilize samples effectively as they discard samples once learned, while their regularization~\cite{TENT,CoTTA} may be too strong to allow learning of new knowledge.

\begin{figure*}[t]
\centering
\includegraphics[width=.85\linewidth]{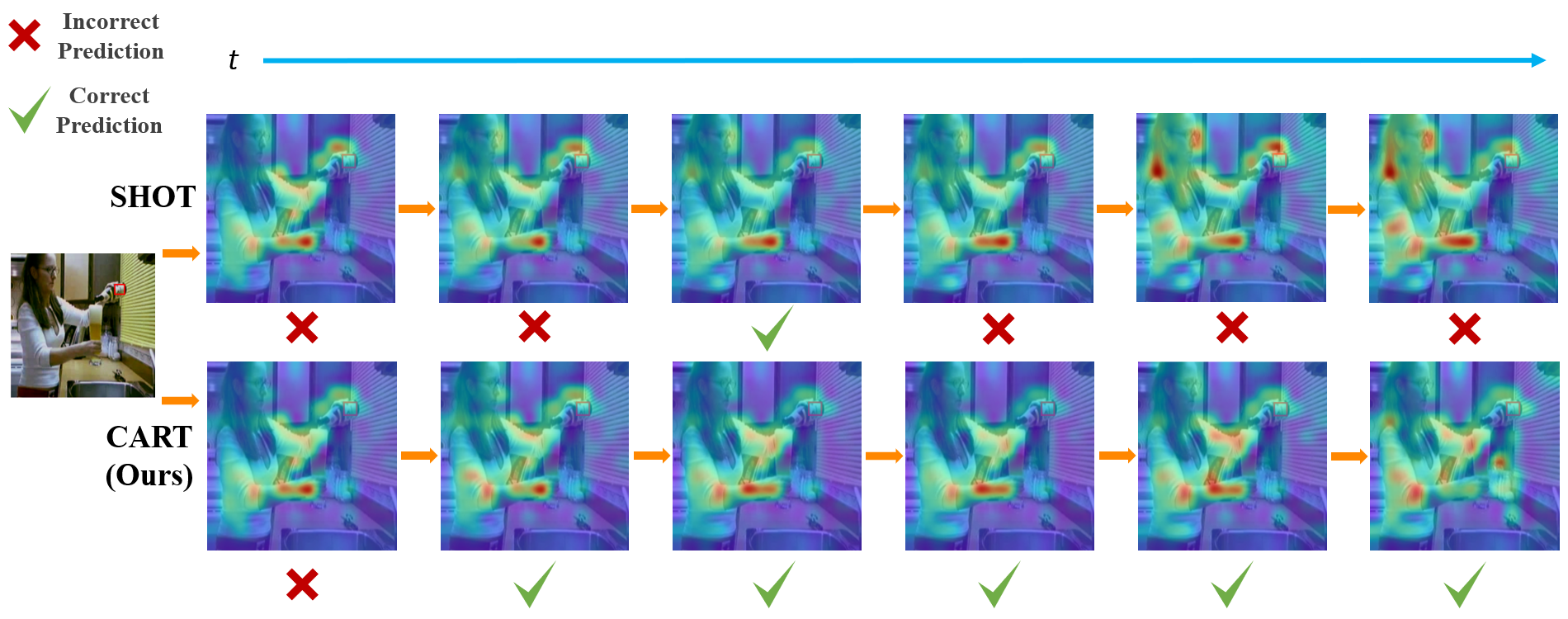}
\caption{\textbf{Saliency maps that show SHOT can gradually mislead the model and result in incorrect predictions (marked by red crosses), while the proposed CART can maintain the focus of the model and enable the model to generate correct predictions (marked by green ticks)}. The shown attention maps are obtained from the first spatial attention head in the 10\textsuperscript{th} layer of the TimeSFormer network using the patch in the red box as a query. Models trained by SHOT and CART are evaluated on the same challenging testing sample, and the saliency maps for the same attention head are displayed above. Warmer colors, e.g., red and orange area, indicate the query is activating keys of those patches, while colder colors, e.g., green and purple, indicate those areas are less attended to by the attention head. \label{exp_fig2}} 
\end{figure*}

\subsection{Ablation Studies}
To analyze the effectiveness of the two modules in CART, we conduct a series of ablation studies on the ARID$\rightarrow$MIT$\rightarrow$HMDB51 sequence and carefully remove each module to investigate their effectiveness. Results are displayed in Table.~\ref{table_5}. First, we evaluate the effectiveness of enhancing distillation with data generalization by removing the generalization, i.e., optimize $\mathcal{L}_{Acls}+\mathcal{L}_{dis}$. Results suggest that the average adaptation performance dropped by 1.60\%, and the average forgetting rate increased by 49.09\%, indicating the model is much more likely to forget previously learned knowledge. Next, we further remove strong augmentation entirely, i.e., optimize $\mathcal{L}_{cls}+\mathcal{L}_{dis}$, and find the adaptation performance is dropped heavily, indicating the accumulation error is less reduced and leading to degraded performance. Lastly, we completely remove distillation in CART, and find that the model performance is decreased again. Overall, while distillation enables a significant reduction of accumulation of errors, the performance can receive another leap when using data generalization.
On the other hand, we are also interested in validating the effectiveness of the proposed attentive learning of pseudo labels. Thus, we set $M=1,\tau=0$ to transform $\mathcal{L}_{Acls}$ into a standard cross-entropy loss. Results reflect that our attentive learning strategy can indeed help to reduce the accumulation of prediction errors and improve model adaptation performance.
Finally, to validate two claims we made in Section \ref{method} that distilling from the model at $t-1$ moment is sub-optimal and simply setting trade-off for distillation high can limit model plasticity, we optimize CART based on $\mathcal{L}_{Acls}+\mathcal{L}_{dis}, \alpha=10$ and $\mathcal{L}_{Acls}+\mathcal{L}_{Adis}'$, respectively. Results from both experiments have sufficiently proven our claims and justify the design of CART.

\begin{table}[t]
\centering
\resizebox{.9\linewidth}{!}{
	\begin{tabular}{c|c|c|c}
		\hline\hline
		method & Loss & Mean Acc. & Mean Forget\\
		\hline
		\multirow{7}{*}{CART} & $\mathcal{L}_{Acls}+\mathcal{L}_{Adis}$ & {\bf 32.55\%} & {\bf -3.83\%}\\
		&  $\mathcal{L}_{Acls}+\mathcal{L}_{dis}$ & 32.01\% & -1.95\% \\
		&  $\mathcal{L}_{cls}+\mathcal{L}_{dis}$ & 30.26\% & -3.28\% \\
		&  $\mathcal{L}_{cls}$ & 28.38\% & 0.60\% \\
		&  $\mathcal{L}_{Acls}+\mathcal{L}_{Adis},m=1, \tau=0$ & 32.16\% & -3.48\% \\
		& $\mathcal{L}_{Acls}+\mathcal{L}_{dis}, \alpha=10$  & 29.49\% & 1.19\%\\
		& $\mathcal{L}_{Acls}+\mathcal{L}_{Adis}'$ &  29.11\%  & 2.57\%\Bstrut\\
		\hline\hline
	\end{tabular}
}
\smallskip
\caption{Ablation study on ARID$\rightarrow$MIT$\rightarrow$HMDB51 \label{table_5}}
\end{table}

\begin{figure}[!t]
\centering
\includegraphics[width=0.9\linewidth]{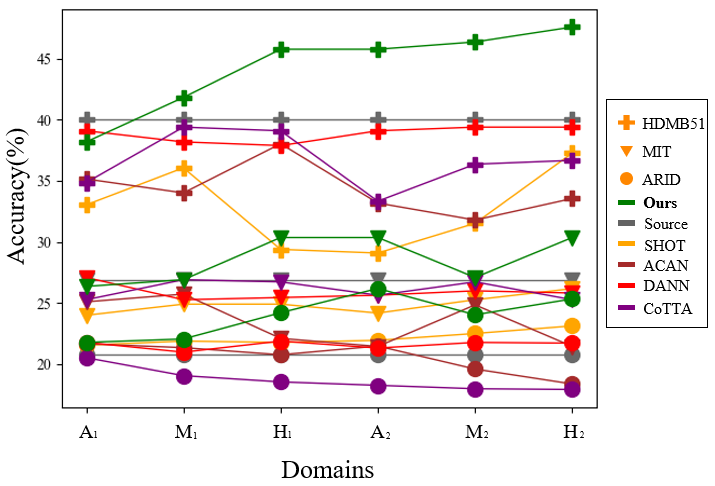}
\caption{Accuracy curves for different methods on the ARID$\rightarrow$MIT$\rightarrow$HMDB51 sequence. 
	A\textsubscript{1}, M\textsubscript{1}, H\textsubscript{1}, A\textsubscript{2}, M\textsubscript{2}, and H\textsubscript{2} denote ARID\textsubscript{1}, MIT\textsubscript{1}, HMDB\textsubscript{1}, ARID\textsubscript{2}, MIT\textsubscript{2}, and HMDB\textsubscript{2}.
	\label{exp_fig1}}
\end{figure}

\subsection{Empirical Analysis}
\noindent\textbf{Adaptation Performance Visualization.} To understand how different methods react to continuous domain shifts, we plot accuracy curves are plotted in Fig.~\ref{exp_fig1}. It also aims to visualize how the reduction of accumulated error is translated into an increase in adaptation performance. We conduct this experiment on ARID$\rightarrow$MIT$\rightarrow$HMDB51 and omit the curve for MCD~\cite{MCD} and TENT~\cite{TENT} because the former has significant performance regression, and the latter is achieving results similar to CoTTA~\cite{CoTTA}. We plot testing results on all domains even if the model has not seen them. According to Fig.~\ref{exp_fig1}, CART is achieving performance gains, particularly on the HMDB51 dataset, while the performances of some other methods are constantly dropping over time. Specifically, we find that most methods struggle to surpass the source model performance, indicating that the accumulated prediction errors are built up over time, and the model forgets learned knowledge. We also found that ARID is particularly challenging as the accuracy of pseudo labels generated on this domain by the source model can be as low as 20\%, making models more likely to be misled when adapting to this domain. In summary, the results of CART on ARID prove that our method can learn from some of the most difficult target domains while also being performative on others, meaning that we had effectively controlled the level of accumulated prediction errors in the learned model. 

\noindent\textbf{Self-Attention Visualization.} Empirically, transformer-based networks rely on attending to a specific object in inputs to perform the extraction of video features. Therefore, we opt to visualize how attention heads evolve as the time step $t$ increases. We obtain saliency maps in Fig.~\ref{exp_fig2}, showing that the attention head is gradually losing its focus on specific objects, e.g., human arms, human body, etc., when SHOT~\cite{SHOT} is employed. In contrast, the saliency maps of CART demonstrate that the model consistently focuses on related objects and evolves to classify this challenging input correctly. In sum, Fig.~\ref{exp_fig2} demonstrates that our novel regularization method can successfully control the detrimental drift of model parameters and thus improve the CVDA performance of modern video models.

\section{Conclusion}
This work proposes a novel Confidence-Attentive network with geneRalization enhanced self-knowledge disTillation (CART) to tackle the new Continuous Video Domain Adaptation (CVDA) task. In CVDA, we assume that target domains are continuously arriving and previously seen domains are unavailable. Without available source data, CART learns from refined pseudo labels attentively to avoid being misled by accumulated prediction errors. CART further reduces the accumulation of prediction errors by deploying a generalization enhanced self-knowledge distillation module. Extensive experiments justify that these two modules can effectively tackle challenges in CVDA, resulting in state-of-the-art adaptation results on dedicated CVDA benchmarks.

\section*{Appendix}
This appendix presents more details about the proposed Confidence-Attentive network with geneRalization enhanced self-knowledge disTillation (CART). We organize this appendix as follows: 
1) we present more benchmark details including learning rates, batchsize, and hyper-parameter settings, about our implementations of CART and other methods, 
2) we present more detailed visualizations of the results achieved by CART to demonstrate that CART enables deep video models to continuously adapt to changing domains with consistently high performances, 
3) we present a detailed description of the datasets we have used, 
and 4) we present a detailed comparison between CART and other related methods to highlight the novelty of CART.

\subsection*{Detailed Implementations}
\noindent \textbf{Brief Review of CART.} In this work, we propose a Confidence-Attentive network with geneRalization enhanced self-knowledge disTillation (CART) to address Continuous Video Domain Adaptation (CVDA) via pseudo labeling, and we mainly focus on reducing the accumulation of prediction errors. CART contains two modules to reduce the accumulation of prediction errors, i.e., attentive learning of pseudo labels and self-knowledge distillation enhanced by data generalization.

\noindent \textbf{Implementation Details of CART.} 
For the distillation process, the augmentation set we used is inspired by RandAugment~\cite{RandAug}. For weak augmentation, we only leverage some common augmentations such as center cropping and normalization~\cite{C3D,TSN,timesformer}. For strong augmentation, we include random HLS color variation, Gaussian noise, camera noise, flipping, etc. We set the distillation temperature to 2 and the trade-off for distillation to 5. For the attentive learning process, we set $m=2,\tau_1=0.1,\tau_2=0.5$. Comprehensively, this means low-confidence samples will be learned once while high-confidence samples are learned twice. For the backbone, we employ TimeSFormer~\cite{timesformer}, which is an efficient and robust transformer-based network. It achieves similar results as bigger models such as Swin~\cite{swin} and ViViT~\cite{vivit}. We load the pre-trained weights trained based on the Kinetics-400~\cite{K400} dataset. For efficiency and accuracy, we freeze the first 4 blocks of the TimeSFormer. We train on each training data split for 10 epochs. The learning rate is set to 0.001 and decreases by 10 times every 5 epochs. For pseudo label generation, CART re-generates the pseudo label every 5 epochs. Stochastic Gradient Descent (SGD)~\cite{SGD} optimizer is used with weight decay set to 0.0001 and momentum set to 0.9. We also leverage nesterov momentum~\cite{nesterov}. We use a batch size of 96 with 16 samples per GPU. Each video clip contains 8 frames sampled from 8 segments of a video.

\begin{figure*}[!t]
	\centering
	\includegraphics[width=.9\linewidth]{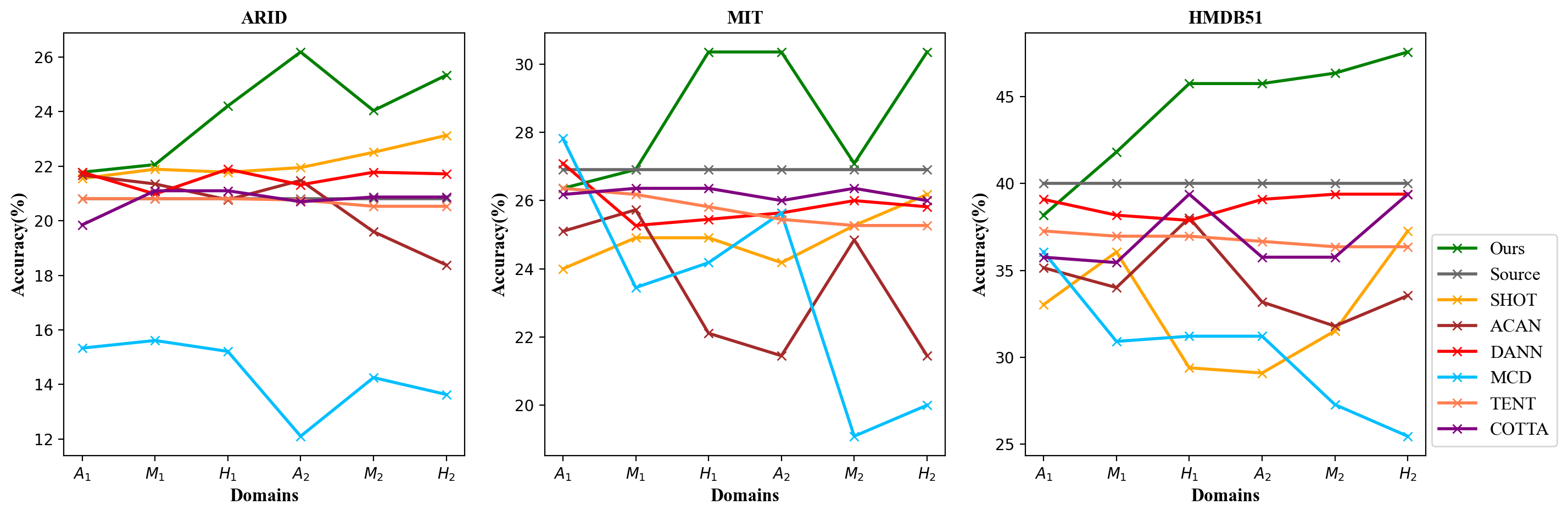}
	\caption{Accuracy Curves on ARID$\rightarrow$MIT$\rightarrow$HMDB51 sequence \label{fig1}} 
\end{figure*}

\begin{figure*}[!t]
	\centering
	\includegraphics[width=.9\linewidth]{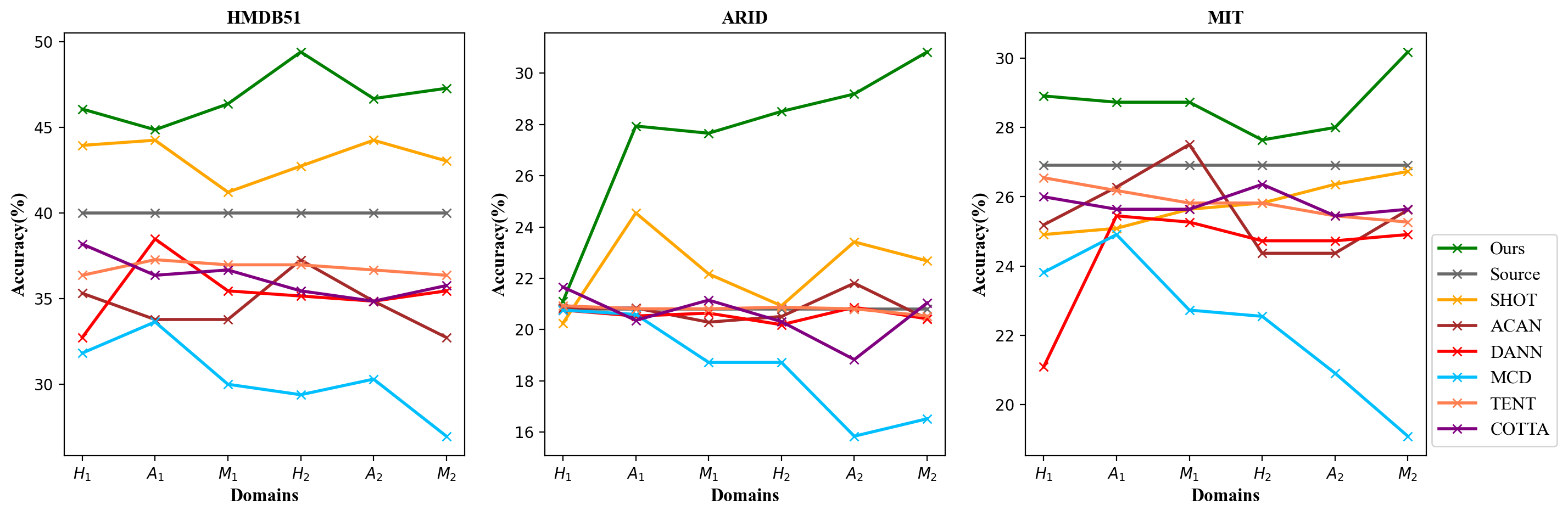}
	\caption{Accuracy Curves on HMDB51$\rightarrow$ARID$\rightarrow$MIT sequence \label{fig2}} 
\end{figure*}

\begin{figure*}[!t]
	\centering
	\includegraphics[width=0.75\linewidth]{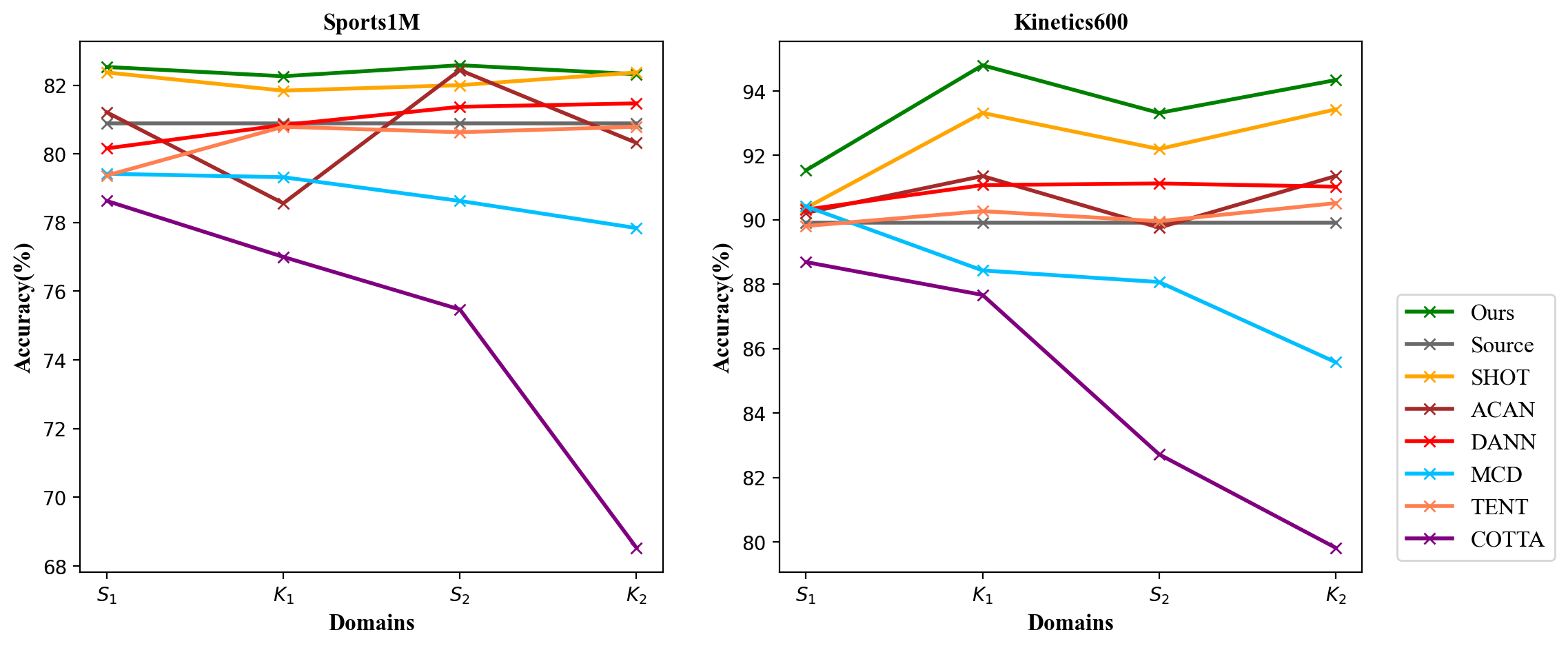}
	\caption{Accuracy Curves on Sports1M$\rightarrow$Kinetics600 sequence \label{fig3}} 
\end{figure*}

\noindent \textbf{Implementation Details of Other Methods} For a fair comparison, we adapt the state-of-the-art methods from related but different research works~\cite{SHOT,DANN,MCD,ACAN,TENT,CoTTA}. SHOT~\cite{SHOT} is directly employed without further modifications as it is compatible with CVDA settings. For UDA~\cite{DANN,MCD} and VUDA~\cite{MCD} methods, we made source data available to them. Instead of feeding 16 samples per GPU, a total of 32 samples (16 sources and 16 targets) are fed to a GPU when UDA/VUDA methods are employed. Since UDA/VUDA methods sometimes encounter stability issues~\cite{stability} in CVDA, we apply gradient clipping to stabilize the training. For TTA methods~\cite{TENT,CoTTA} the training and testing split is combined as one data stream. We iterate the model directly on this data stream and the model will randomly encounter either a training or a testing sample, though only the accuracy of the testing samples will be recorded.

\begin{figure*}[!t]
	\centering
	\includegraphics[width=.9\linewidth]{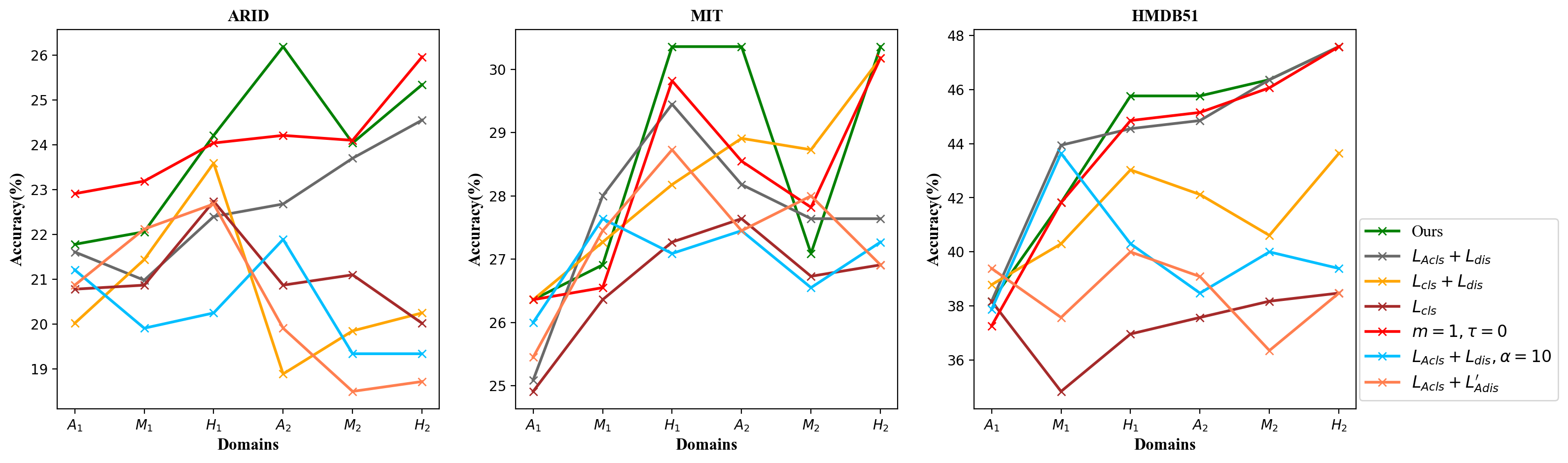}
	\caption{Accuracy Curves for Ablation Studies \label{fig4}} 
\end{figure*}

\begin{figure*}[!t]
	\centering
	\includegraphics[width=.8\linewidth]{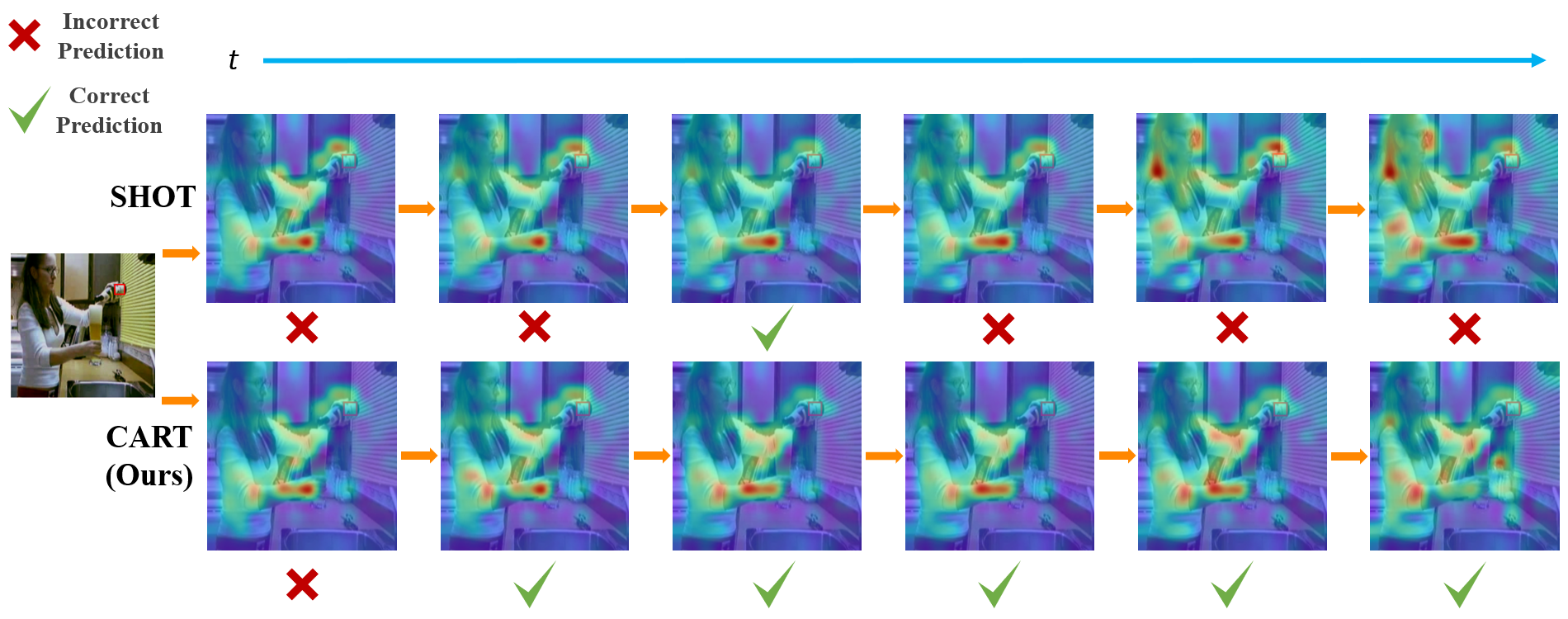}
	\caption{Saliency map visualization of a spatial attention head in TimeSFormer. Red crosses mean the prediction is incorrect, and green ticks mean the prediction is correct. The action performed in the video is \textit{'pouring'}, and we set the query near the glass bottle (red bounding box). The resulting visualization show that CART can constantly maintain the focus of the model while SHOT is likely to be misled by accumulated errors and the focus on relevant objects is losing gradually. This is also reflected by the fact that CART is learning to correctly classify this sample while SHOT is achieving unstable results. Warmer colors indicate greater attention, while colder colors indicate lower attention on those patches.\label{fig5}} 
\end{figure*}

\begin{figure*}[!t]
	\centering
	\includegraphics[width=.8\linewidth]{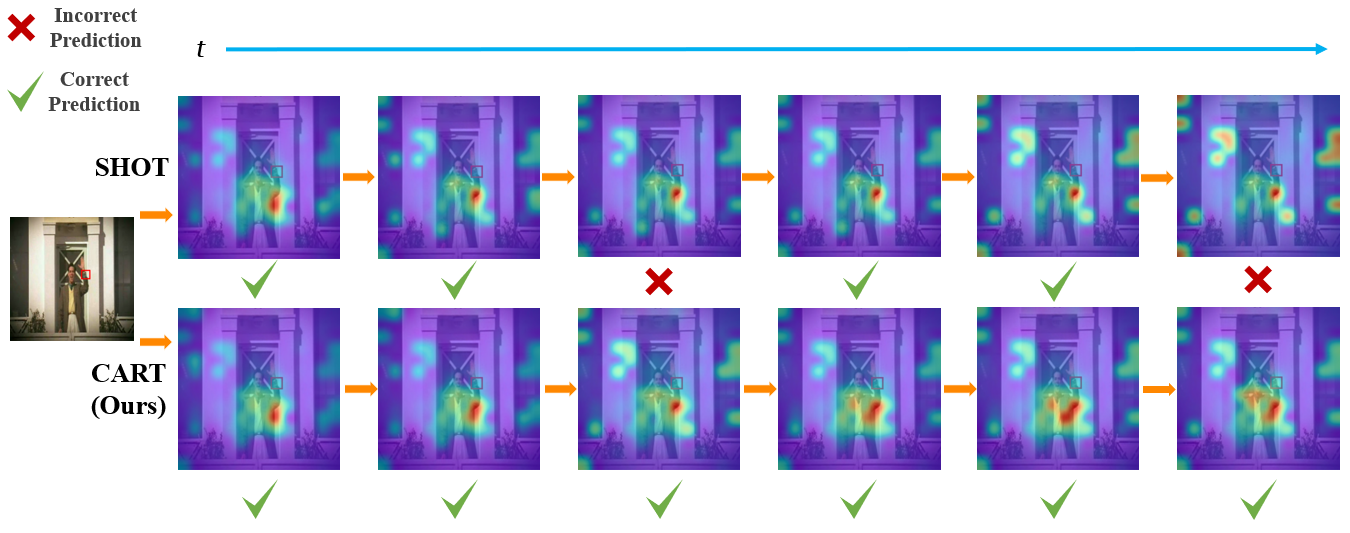}
	\caption{Saliency map visualization of a spatial attention head in TimeSFormer. Red crosses mean the prediction is incorrect, and green ticks mean the prediction is correct. The action performed in the video is \textit{'waving'}. The query is set to near the man's arm (red bounding box). The results suggest that the model adapted by SHOT is shifting its focus onto irrelevant backgrounds while the model adapted by CART can attend to the human body even more. We also find that the model adapted by SHOT is likely to make incorrect predictions when it encounters the MIT dataset. Warmer colors indicate greater attention, while colder colors indicate lower attention on those patches.\label{fig6}} 
\end{figure*}

\begin{table*}[!t]
	\center
	\resizebox{1\linewidth}{!}{
		\smallskip\begin{tabular}{c|c|c|c|c}
			\hline
			\hline
			Class ID & ARID Class & HMDB51 Class & Moments-in-Time Class & Kinetics-600 Class\\
			\hline
			0 & Drink & drink & drinking & drinking shots \\
			1 & Jump & jump & jumping & jumping bicycle, jumping into pool, jumping jacks\\
			2 & Pick & pick & picking & picking fruit\\
			3 & Pour & pour & pouring & pouring beer\\
			4 & Push & push & pushing & pushing car, pushing cart, pushing wheelbarrow, pushing wheelchair\\
			5 & Run & run & running & running on treadmill\\
			6 & Walk & walk & walking & walking the dog, walking through snow\\
			7 & Wave & wave & waving & waving hand \\
			8 & Sit & sit & sitting & falling off chair\\
			9 & Stand & stand & standing & snatch weight lifting\\
			10 & Turn & turn & turning & pirouetting \\
			\hline
			\hline
		\end{tabular}
	}
	\smallskip
	\caption{List of action classes for Daily-DA\textsubscript{Conti.}. \label{supp_table_2}}
\end{table*}

\begin{table*}[!ht]
	\center
	\resizebox{.85\linewidth}{!}{
		\begin{tabular}{c|c|c|c}
			\hline
			\hline
			Class ID & UCF101 Class & Sports-1M Class & Kinetics-600 Class\\
			\hline
			0 & Archery & archery & archery \Tstrut\\
			1 & Baseball Pitch & baseball & catching or throwing baseball, hitting baseball\\
			2 & Basketball Shooting & basketball & playing basketball, shooting basketball\\
			3 & Biking & bicycle & riding a bike\\
			4 & Bowling & bowling & bowling\\
			5 & Breaststroke & breaststroke & swimming breast stroke\\
			6 & Diving & diving & springboard diving\\
			7 & Fencing & fencing & fencing (sport)\\
			8 & Field Hockey Penalty & field hockey & playing field hockey\\
			9 & Floor Gymnastics & floor (gymnastics) & gymnastics tumbling\\
			10 & Golf Swing & golf & golf chipping, golf driving, golf putting\\
			11 & Horse Race & horse racing & riding or walking with horse\\
			12 & Kayaking & kayaking & canoeing or kayaking\\
			13 & Rock Climbing Indoor & rock climbing & rock climbing\\
			14 & Rope Climbing & rope climbing & climbing a rope\\
			15 & Skate Boarding & skateboarding & skateboarding\\
			16 & Skiing & skiing & skiing crosscountry, skiing mono\\
			17 & Sumo Wrestling & sumo & wrestling\\
			18 & Surfing & surfing & surfing water\\
			19 & Tai Chi & t'ai chi ch'uan & tai chi\\
			20 & Tennis Swing & tennis & playing tennis\\
			21 & Trampoline Jumping & trampolining & bouncing on trampoline\\
			22 & Volleyball Spiking & volleyball & playing volleyball \Bstrut\\
			\hline
			\hline
		\end{tabular}
	}
	\smallskip
	\caption{List of action classes for Sports-DA\textsubscript{Conti.}. \label{supp_table_3} }
\end{table*}

\subsection*{Visualizations of Continuous Video Domain Adaptation}
In this section, we offer a better view of the accuracy curves of all methods on all benchmarks we have tested on, including the curves for experiments in ablation studies and additional saliency map visualizations~\cite{vit}. In Fig.~\ref{fig1},\ref{fig2},\ref{fig3},\ref{fig4}, we show accuracy curves on each dataset (domain), individually in three subplots for better clarity. Specifically, in Fig.~\ref{fig1}, the results from ARID$\rightarrow$MIT$\rightarrow$HMDB51 are displayed and CART achieves an average relative improvement of 10.05\% over the previous state-of-the-art method. In Fig.~\ref{fig2},\ref{fig3}, similarly, CART is also constantly achieving relative improvements of 12.78\% and 0.83\% over previous state-of-the-art methods, respectively. In Fig.~\ref{fig4}, the ablation studies are showing that the proposed attentive learning of refined pseudo labels and generalization-enhanced self-knowledge distillation can enable deep video models to effectively adapt to continuously changing target domains. For the saliency maps in Fig.~\ref{fig5},\ref{fig6}, the raw attention maps are obtained from the output of the first spatial attention head in the 10\textsuperscript{th} layer of the TimeSFormer network and visualized following the standard practice~\cite{vit}, and they have clearly demonstrated that the proposed CART can prevent the model from being misled by accumulated errors. In summary, all four figures have effectively proven that CART is constantly achieving state-of-the-art CVDA results.

\subsection*{CVDA Benchmarks}
In this paper, we closely simulate real-world CVDA scenarios by modifying the Daily-DA and Sports-DA datasets proposed in~\cite{MSVDA}. Shared classes are listed in Table.~\ref{supp_table_2},~\ref{supp_table_3}. Statistics of both Daily-DA\textsubscript{Conti.} and Sports-DA\textsubscript{Conti.} are listed in Table~\ref{supp_table_1}.

\begin{table*}[!ht]
	\center
	\resizebox{.8\linewidth}{!}{
		\begin{tabular}{c|c|c}
			\hline
			\hline
			Statistics & Daily-DA & Sports-DA  \Tstrut\Bstrut\\
			\hline
			Video Classes \# & 11 & 23 \Tstrut\\
			Training Video \# & A:3,792 / H:770 / M:5,500 / K600:10,639 & U:2,145 / S:14,754 / K600:19,104\\
			Testing Video \# & A:1,768 / H:330 / M:550 / K:725 & U:851 / S:1,900 / K:1,961 \Bstrut\\
			\hline
			\hline
		\end{tabular}
	}
	\smallskip
	\caption{Statics of Daily-DA\textsubscript{Conti.} and Sports-DA\textsubscript{Conti.} A, H, M, and K600 refer to ARID, HMDB51, Moments-in-Time, and Kinetics-600. Notice that these two datasets use a different subset of Kinetics-600.}
	\label{supp_table_1}
\end{table*}

\begin{table*}[!ht]
	\center
	\resizebox{.95\linewidth}{!}{
		\begin{tabular}{m{.14\textwidth}|m{.1\textwidth}|m{.35\textwidth}|m{.45\textwidth}}
			\hline
			\hline
			Method & Publication & Task & Techniques \Tstrut\Bstrut\\
			\hline
			SHOT~\cite{SHOT} & ICML-20 & Source-Free Domain Adaptation (SFDA): source data is unavailable for adaptation, and target data is available without supervision. & (a) SHOT prepares a source model with labeling smoothing loss; (b) SHOT trains a model with Information-Maximization loss~\cite{IMLoss}; (c) SHOT re-classifies model predictions to the closest prototype of each prediction multiple times~\cite{deepcluster}\\
			\hline
			DANN~\cite{DANN} & JMLR-16 & Unsupervised Domain Adaptation (UDA): source data is unavailable for training, and target data is unsupervised. & (a) DANN performs classification similar to other networks; (b) DANN additionally introduces a domain adversarial network; (c) the domain adversarial network tries to classify samples to source and target domain while the feature extractor tries to confuse the domain adversarial network.\\
			\hline
			ACAN~\cite{ACAN} & TNNLS-22 & Video-based Unsupervised Domain Adaptation (VUDA): Source video data is available and supervised, and target video data is available without supervision. & (a) ACAN is based on adversarial domain adaptation across the spatial-temporal video features; (b) it aligns video correlation features in the form of long-range spatial-temporal dependencies; (c) it aligns the joint distribution of correlation information of different domains by minimizing pixel correlation discrepancy.\\
			\hline
			CoTTA~\cite{CoTTA} & CVPR-22 & Continuous Test-Time Domain Adaptation (CoTTA): the test data stream is directly evaluated and the model trains itself online using only the test data. & (a) CoTTA uses a weighted-averaged model to generate pseudo labels; (b) CoTTA applies augmentation to samples whose confidence is low; (c) CoTTA randomly restore a small portion of the model parameter back to their initial state in the source model.\\
			\hline
			\textbf{CART(Ours)} & - & Continuous Video Domain Adaptation (CVDA): unsupervised and changing target domain comes sequentially, source data and previous target domain data are not available. & (a) CART generates refined pseudo labels for the arriving target domain; (b) CART learns from unreliable pseudo labels attentively to avoid being misled and reduce the accumulation of errors; (c) CART additionally learns from the source model and regularizes the model by keeping the output of strongly augmented samples similar to outputs of weakly augmented samples from the source model.\\
			\hline
			\hline
		\end{tabular}
	}
	\smallskip
	\caption{Detailed comparison of CART with related but different SFDA, UDA/VUDA, and TTA methods.}
	\label{supp_table_4}
\end{table*}

Compared to the original Daily-DA, we introduced 3 additional classes into Daily-DA to further extend it, i.e., sitting, standing, and turning. Therefore, there are a total of 11 classes in Daily-DA\textsubscript{Conti.}. Among them, ARID~\cite{ARID} is a particularly challenging dataset as it contains videos shot in a dark environment, while the other three~\cite{K600,moments-in-time,HMDB51} contain videos shot in normal lighting conditions. This makes domain adaptation to ARID more challenging than others. Moreover, Moments-in-Time~\cite{moments-in-time} is also different from Kinetics-600~\cite{K600} and HMDB51~\cite{HMDB51} because it contains non-human actions. For instance, the action can be performed by a cartoon character or an animal.

For Sports-DA\textsubscript{Conti.}, it is directly adapted from its original version proposed in \cite{MSVDA}. A total of 23 classes are involved, and the domain shift between each dataset in this benchmark is much smaller, making it easier for a pre-trained model to adapt to this domain. Therefore, we choose UCF101~\cite{UCF101} as the source domain as the source model obtained from this domain has a lower generalization ability, making it a suitable benchmark for checking the effectiveness of domain adaptation methods.

To simulate real-world scenarios where the seen domains can re-appear during long-time continuous adaptation, we further split each target dataset into two non-overlapping subsets randomly. The model first iterates through all target domains once and then repeats the sequence using another unseen split. We admit that there can be many other possible data splits and training orders. Nevertheless, our benchmarks mainly focus on two sequences, i.e., ARID$\rightarrow$MIT$\rightarrow$HMDB51 and HMDB51$\rightarrow$ARID$\rightarrow$MIT on Daily-DA\textsubscript{Conti.} and Sports1M$\rightarrow$Kinetics600 on Sport-DA\textsubscript{Conti.} because those sequences are representative and can sufficiently and effectively demonstrate the CVDA performance of involved methods.

\subsection*{Detailed Comparison with Related but Different Methods}
\label{section:supp:detail-compare}
In this paper, we proposed CART to address the challenges underlying the CVDA scenario. It achieves state-of-the-art results on Daily-DA\textsubscript{Conti.} and Sports-DA\textsubscript{Conti.}. An average of 8.42\% relative improvement is observed on two benchmarks, demonstrating that our method can effectively adapt to continuously changing domains. Moreover, CART is also consistently retaining its knowledge learned from previously seen domains, and this is reflected by a -1.10\% average forgetting rate (the model is learning instead of forgetting). We highlight our novelty by comparing it with related works that can partially tackle the underlying challenges of the CVDA scenario. Specifically, we compare CART to SFDA~\cite{SHOT} methods, UDA/VUDA methods~\cite{DANN,MCD,ACAN}, and TTA methods~\cite{TENT,CoTTA}. These methods are compared in Table~\ref{supp_table_4} from two perspectives: the task they tackle and the techniques they proposed.

{\small
\bibliographystyle{ieee_fullname}
\bibliography{egbib}
}

\end{document}